\documentclass[conference]{IEEEtran}
\IEEEoverridecommandlockouts
\usepackage{cite}
\usepackage{amsmath,amssymb,amsfonts}
\usepackage{algorithmic}
\usepackage{graphicx}
\usepackage{textcomp}
\usepackage{xcolor}
\usepackage{bm}
\usepackage{booktabs}
\usepackage{capt-of}
\def\BibTeX{{\rm B\kern-.05em{\sc i\kern-.025em b}\kern-.08em
    T\kern-.1667em\lower.7ex\hbox{E}\kern-.125emX}}
\begin{document}

\title{FAN: Foresight Action Normalization for Continual Adaptation of Vision-Language-Action Models\\
}

\author{
Yijun Hong\textsuperscript{1,4,*}, Jiarun Zhu\textsuperscript{3,*}, Xiaoquan Sun\textsuperscript{5,*}, Le Xu\textsuperscript{1}, Qijun He\textsuperscript{1}, Xin Jin\textsuperscript{3}, Mingqi Yuan\textsuperscript{1,2,$\dagger$},\\ Wenjun Zeng\textsuperscript{3,$\ddagger$}, Jiayu Chen\textsuperscript{1,2,$\ddagger$}

\\
\textsuperscript{1}HKU\quad\textsuperscript{2}INFIFORCE\quad\textsuperscript{3}EIT, Ningbo\quad\textsuperscript{4}SUSTech\quad\textsuperscript{5}HUST
\thanks{$*\:$These authors contributed equally.}
\thanks{$\dagger\:$Project lead: Mingqi Yuan ({\tt\small my017@hku.hk}).}
\thanks{$\ddagger\,$Corresponding authors: Wenjun Zeng (\texttt{\small wzeng-vp\allowbreak @eitech\allowbreak .edu.cn}) and Jiayu Chen (\texttt{\small jiayuc\allowbreak @hku\allowbreak .hk}).}
}


\IEEEaftertitletext{%
    \begin{minipage}{\textwidth}
        \centering
        \includegraphics[width=\linewidth]{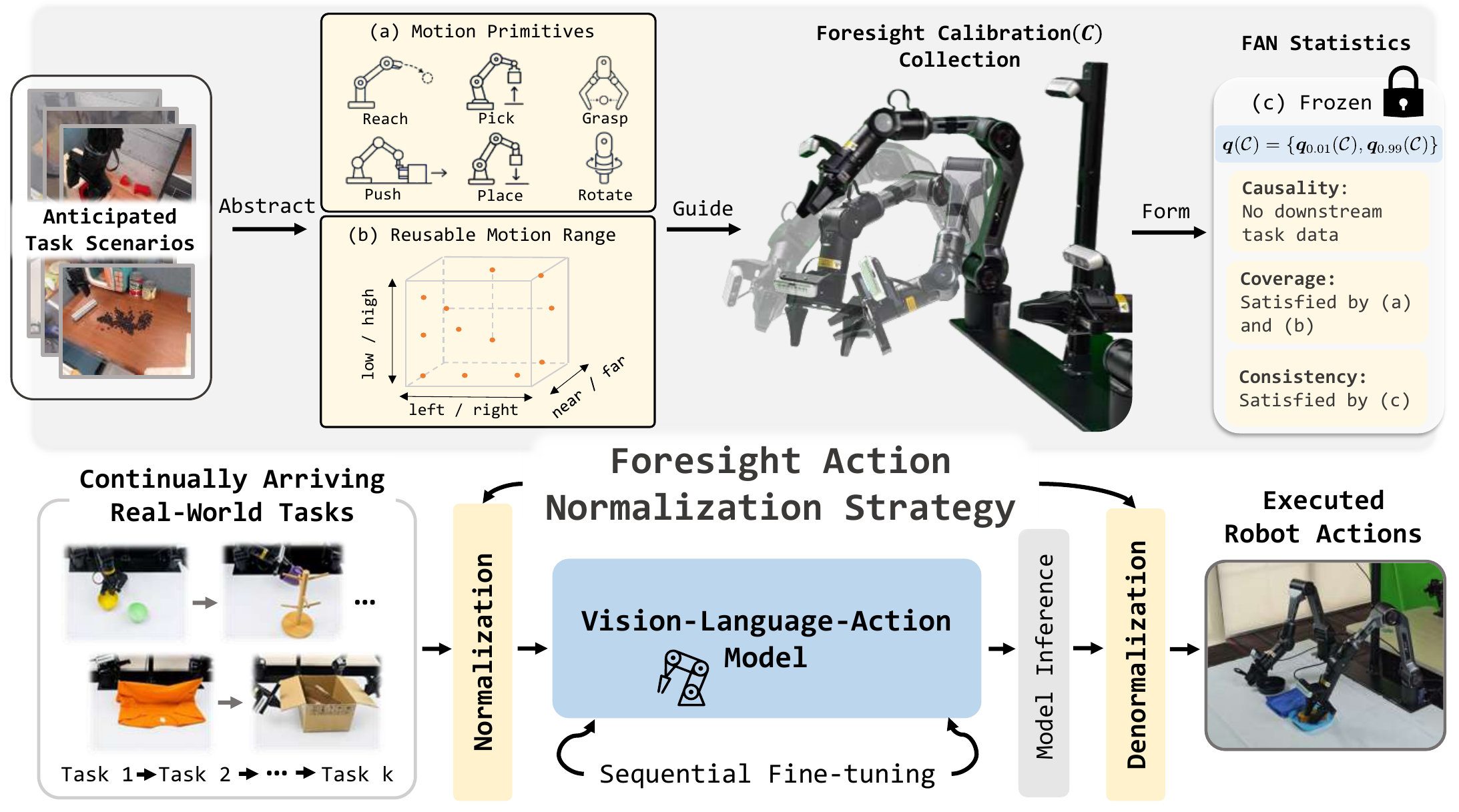}
        \captionof{figure}{
        Overview of Foresight Action Normalization (FAN). Before the continual task stream, coarse anticipated deployment scenarios and the target embodiment's capabilities are manually abstracted into (a) motion primitives and (b) a reusable embodiment-level motion range. These motion requirements guide the collection of a task-independent calibration set $\mathcal{C}$. (c) The resulting percentile statistics, $\bm{q}(\mathcal{C})=\{\bm{q}_{0.01}(\mathcal{C}), \bm{q}_{0.99}(\mathcal{C})\}$, are estimated once and frozen for normalization and denormalization throughout sequential fine-tuning and inference. FAN preserves causality by excluding future-task data, supports coverage through (a) and (b), and ensures consistency by freezing (c),  providing a fixed, task-independent action coordinate system across tasks.
        }
        \label{fig:preface}
    \end{minipage}
}

\maketitle

\begin{abstract}
Vision-Language-Action (VLA) models pre-trained on large-scale, closed datasets have demonstrated remarkable success across diverse robotic manipulation tasks. However, their long-term real-world deployment necessitates continuously acquiring new skills while retaining previously learned capabilities. While pioneering works have explored continual VLA adaptation using techniques such as experience replay and reinforcement fine-tuning, they overlook a foundational mechanism: action normalization, which determines the underlying coordinate system in which policies perceive and execute physical actions. To bridge this gap, we systematically evaluate five normalization strategies across four real-world task streams covering single-arm and bimanual manipulation. Our analysis reveals that existing protocols induce severe failure modes due to inter-task coordinate drift, limited motion coverage, or train-test coordinate mismatches. Motivated by these insights, we formulate three core design principles: consistency, coverage, and causality (3C), and introduce foresight action normalization (FAN). FAN estimates normalization statistics once from a small, task-independent calibration set prior to continual learning and freezes them throughout adaptation. Across all evaluated streams, FAN achieves the highest performance and demonstrates consistent robustness, providing insightful guidance for building stable action representations in achieving effective lifelong VLA adaptation.
\end{abstract}


\section{Introduction}
A general-purpose robotic agent must perform a broad range of tasks while continually expanding its skill repertoire through ongoing interaction with the world. Vision-Language-Action (VLA) models provide a promising foundation for broad task competence by integrating visual perception, language understanding, and action prediction within a unified policy \cite{kim2024openvla,black2025pi0,black2025pi05}. Recent systems further extend these capabilities through cross-embodiment soft prompting, lightweight deployment-oriented designs, and interleaved visual-language instructions \cite{zheng2026xvla,song2026llavavla,fan2026interleavevla}. However, broad competence acquired through static pretraining alone is insufficient for reliable adaptation after deployment, where robots inevitably encounter unexpected tasks and changing physical dynamics. This necessitates a shift from static policy execution toward continual adaptation, in which VLA models sequentially acquire new skills while avoiding catastrophic forgetting of previously learned capabilities \cite{wang2024comprehensive}. Establishing effective mechanisms for such adaptation remains a fundamental requirement for scalable, lifelong robotic intelligence.

To mitigate catastrophic forgetting during real-world adaptation, recent studies on continual VLA learning have explored techniques such as experience replay \cite{rolnick2019experience,liu2026continualvla,zhu2026continualvla,chen2026phaser}, active recovery sampling \cite{karli2026recall}, adapter expansion \cite{romer2026clare}, and post-training optimization \cite{liu2026longlived}. While these efforts primarily concentrate on optimization objectives, architectural expansion, or memory management, they universally overlook a foundational component inherited from offline pipelines: \textbf{action normalization}. In continuous control, normalization statistics define the coordinate system in which a policy represents and predicts actions. 

This coordinate system becomes particularly important during continual adaptation, where a single policy must learn from heterogeneous tasks with different motion ranges, kinematic configurations, and interaction dynamics. Although normalization is often treated as a preprocessing detail, we find that its design can fundamentally determine whether continual adaptation succeeds or fails.

In this work, we present the first systematic study of action normalization strategies in the continual adaptation of VLA models. Through extensive real-robot evaluations across single-arm and bimanual task streams, we reveal that prevailing normalization protocols produce distinct failures in skill acquisition, transfer, and retention. 
These empirical findings, together with the unavailability of future-task data in continual imitation learning, motivate three essential design principles for continual normalization: Consistency, Coverage, and Causality (3C).
Motivated by the 3C principle, we propose a simple yet remarkably effective strategy entitled \textbf{F}oresight \textbf{A}ction \textbf{N}ormalization (\textbf{FAN}). As illustrated in Fig.~\ref{fig:preface}, FAN estimates normalization statistics once from a small, task-independent calibration set collected before the task stream and freezes them throughout sequential fine-tuning and deployment. Across diverse evaluation streams, FAN achieves the highest performance with exceptional task-order robustness. Our contributions are threefold:
\begin{itemize}
    \item We identify action normalization as a critical yet overlooked factor in continual VLA adaptation and systematically characterize the failure modes of four representative normalization protocols.
    \item We formulate Consistency, Coverage, and Causality (3C) as foundational design principles for action normalization in continual VLA adaptation.
    \item We introduce FAN, a one-time calibration-based normalization strategy, and validate its efficacy and robustness through four manipulation task streams spanning single-arm and bimanual configurations.
\end{itemize}

\section{Related Work}

\subsection{Vision-Language-Action Models}

Vision-Language-Action (VLA) models integrate visual perception,
language conditioning, and action prediction within a unified robot policy.  OpenVLA and the $\pi$ family
combine large-scale vision-language pretraining, heterogeneous robot data, and expressive action decoders\cite{kim2024openvla,black2025pi0,black2025pi05}. Recent systems extend this paradigm with embodied reasoning and richer conditioning \cite{gemini2025robotics15,ai2026pi07}, scalable cross-embodiment learning\cite{zheng2026xvla,luo2026beingh05}, and more practical or flexible policy interfaces \cite{song2026llavavla,fan2026interleavevla}. Long-horizon robustness and memory are also emerging priorities \cite{sun2026atomvla,dai2026robomme}. 
Whereas these advances broaden VLA capabilities and deployment scope, we study how normalization statistics should be defined and maintained for pretrained VLAs learning from sequential real-world task streams.

\subsection{Continual Learning for VLA Models}

In continual VLA adaptation, a pretrained policy must acquire new tasks sequentially while retaining previously learned behaviors. Data-centric work uses experience replay~\cite{liu2026continualvla,zhu2026continualvla}, phase-aware and semantic replay selection~\cite{chen2026phaser}, or active collection of recovery data~\cite{karli2026recall} to balance acquisition and retention. Model- and optimization-centric approaches instead use parameter-efficient adaptation, autonomous adapter expansion, reinforcement fine-tuning, or continual policy distillation~\cite{zeng2026crlvla,hu2026continualrl,he2026lifelongvla,romer2026clare,liu2026longlived,li2026continualdistillation}. Across these approaches, action normalization is generally inherited from offline VLA pipelines rather than studied as a continual-learning design choice. Yet continual learning precludes using future task data, while updating statistics across stages changes the numerical interface inherited by the policy. Their joint implications for normalization remain largely unexplored in continual VLA adaptation.

\subsection{Action Representation and Normalization in VLA Models}

VLA models employ diverse action representations across datasets and architectures \cite{kawaharazuka2025vlasurvey,zhong2025actionsurvey}. Commands may use joint or end-effector coordinates and absolute or relative actions, while policies employ discrete tokens, continuous regression, diffusion, or flow matching \cite{openx2024,kim2024openvla,black2025pi0}. Recent work further shows that proprioceptive interfaces, observation--action geometric alignment, and cross-embodiment control spaces materially affect policy learning \cite{zhao2026proprioception,chen2026oasis,zheng2026xvla,luo2026beingh05}. 
Beyond numerical action parameterizations, prior work uses changes in hand--object and object--object contact relations to segment and organize manipulation actions \cite{worgotter2013ontology}, while mechanics-based taxonomies group motions by execution-relevant attributes such as contact, trajectory, and arm participation \cite{paulius2019taxonomy}. These abstractions provide a principled basis for translating high-level deployment scenarios into reusable motion requirements.
Continuous states and actions are commonly normalized using moments, bounds, or robust percentiles; OpenVLA and the $\pi$ series, for example, use percentile-based statistics \cite{kim2024openvla,black2025pi0}. Moreover, changing unnormalization metadata while holding model weights fixed can alter the executed physical policy \cite{tai2026sameweights}. Nevertheless, normalization statistics are generally treated as fixed dataset metadata rather than a design choice for continual task streams, which is the gap addressed in this work.

\section{PRELIMINARIES}
\label{sec:preliminaries}

\subsection{Continual Imitation Learning}
\label{subsec:continual_il}

We consider a sequence of robot manipulation tasks
$\{T_k\}_{k=1}^{K}$. Each task is modeled as a finite-horizon
Markov decision process (MDP)~\cite{puterman1990markov},
$\mathcal{M}_k=(\mathcal{S},\mathcal{A},\mathcal{T},
H,\mu_{0,k},R_k)$, where $\mathcal{S}$ and $\mathcal{A}$ are
the state and action spaces,
$\mathcal{T}:\mathcal{S}\times\mathcal{A}\rightarrow\mathcal{S}$
is the transition function, $H$ is the episode horizon and
$\mu_{0,k}$ is the initial-state distribution. All tasks share
$(\mathcal{S},\mathcal{A},\mathcal{T},H)$ but may differ in
$(\mu_{0,k},R_k)$.

A single task-conditioned policy
$\pi_\theta(\bm{a}_t\mid\bm{o}_{\leq t};T_k)$ is learned from
the expert demonstration dataset
$\mathcal{D}_k=\{\tau_i^{(k)}\}_{i=1}^{N_k}$ provided for each
task. Each trajectory $\tau_i^{(k)}=\{(\bm{o}_t,\bm{a}_t)\}_{t=0}^{l_i^{(k)}}$, with
$l_i^{(k)}\leq H$, contains multimodal observations and continuous robot actions, and $\bm{o}_{\leq t}\triangleq(\bm{o}_0,\ldots,\bm{o}_t)$ denotes the observation history available to the policy. If demonstrations from all tasks observed through stage $k$ were retained, the joint
behavior-cloning objective would be
\begin{equation}
\mathcal{L}^{\mathrm{CIL}}_k(\theta)
=
\frac{1}{k}\sum_{p=1}^{k}
\mathbb{E}_{(\bm{o}_{\leq t},\bm{a}_t)\sim\mathcal{D}_p}
\left[
\ell\!\left(
\pi_\theta(\cdot\mid\bm{o}_{\leq t};T_p),
\bm{a}_t
\right)
\right],
\label{eq:cil_objective}
\end{equation}
where $\ell$ denotes the per-sample VLA action-prediction loss, and
$\{\mathcal{D}_{p}:p<k\}$ is not fully available when learning $T_k$.

\subsection{Action Normalization in Continual VLA Learning}
\label{subsec:action_normalization}
Let $\bm{a}_t \in \mathbb{R}^{d_n}$ denote a robot action
vector in physical units. For a reference dataset $\mathcal{D}$, let
$\bm{q}_{\alpha}(\mathcal{D})
=
[q_{\alpha,1}(\mathcal{D}),\ldots,q_{\alpha,d_n}(\mathcal{D})]^\top$
denote the vector of per-dimension action quantiles at level
$\alpha\in\{0.01,0.99\}$, and define
$\bm{q}(\mathcal{D})
=
\{
\bm{q}_{0.01}(\mathcal{D}),
\bm{q}_{0.99}(\mathcal{D})
\}$.
For each action dimension $d\in\{1,\ldots,d_n\}$, normalization is defined as
\begin{equation}
\tilde{a}_{t,d}
=
2
\frac{
a_{t,d}-q_{0.01,d}(\mathcal{D})
}{
q_{0.99,d}(\mathcal{D})-q_{0.01,d}(\mathcal{D})+\epsilon
}
-1,
\label{eq:quantile_normalization}
\end{equation}
where $\epsilon>0$ is a fixed numerical-stability constant.
Inference applies the corresponding inverse transformation,
\begin{equation}
\begin{aligned}
a_{t,d}
=
q_{0.01,d}(\mathcal{D})
+
\frac{\tilde{a}_{t,d}+1}{2}
\left[
q_{0.99,d}(\mathcal{D})-q_{0.01,d}(\mathcal{D})+\epsilon
\right].
\label{eq:quantile_denormalization}
\end{aligned}
\end{equation}

For every strategy considered below, state and action statistics are estimated separately but follow the same provenance and update schedule. We formulate the action transformation explicitly because it determines the numerical interface between the policy output and the physical robot command. Accordingly, the experiments compare complete state and action normalization protocols rather than action-only interventions.

Let $\mathcal{F}_\mathcal{D}$ and $\mathcal{F}_\mathcal{D}^{-1}$ denote the
component-wise normalization and denormalization mappings defined by Eqs.~(2) and~(3), respectively. Equivalently,
$\tilde{a}_t=\mathcal{F}_\mathcal{D}(a_t)$ and
$a_t=\mathcal{F}_\mathcal{D}^{-1}(\tilde{a}_t)$. The reference statistics therefore define the action coordinate
system of the policy. 
For two different reference datasets $\mathcal{D}$ and
$\mathcal{D}'$,
\begin{equation}
\mathcal{F}_\mathcal{D}(a) \neq \mathcal{F}_\mathcal{D'}(a),
\qquad
\mathcal{F}_\mathcal{D}^{-1}(\tilde{a})
\neq
\mathcal{F}_\mathcal{D'}^{-1}(\tilde{a}).
\end{equation}
Thus, in continual learning, changing the statistics alters both the normalized representation of a physical action and the physical interpretation of a prediction. 

\section{Foresight Action Normalization}
\label{sec:method}

A continual normalization protocol should use statistics that are available without future-task data, cover anticipated motions, and remain stable across stages. These considerations motivate three design principles and our calibration-based strategy, FAN.

\subsection{Design Principles for Continual Normalization}
\label{subsec:3c}

\paragraph{Consistency}
The numerical meaning of each normalized action dimension
should remain fixed across stages and across current-task
training, replay, and evaluation. Changing the statistics while
retaining the policy parameters maps the same physical action
to a different normalized value, causing inter-stage
coordinate drift.

\paragraph{Coverage}
The normalization range should cover the embodiment-level
motions anticipated for deployment rather than only the
distribution of a single task. Excessive width should likewise
be avoided because it compresses task-relevant variation in
normalized coordinates. Coverage is therefore a range-matching
requirement rather than a range-maximization objective.

\paragraph{Causality}
The normalization statistics must be available before adaptation and must not depend on future-task data. Deployment should not require external selection of task-specific statistics, since this assumes knowledge of the task identity outside the policy.

These principles expose a fundamental tension: (i) Task-specific statistics provide good local scaling but preclude a shared coordinate system; (ii) statistics fixed from an early task preserve consistency but may not cover later motions; and (iii) dynamically recomputed statistics improve coverage but shift the coordinates inherited by the policy. The objective is therefore to establish an appropriately scoped range before continual adaptation and keep the statistics fixed thereafter.

\subsection{Foresight Action Normalization}
\label{subsec:fan}

FAN implements the above principles by decoupling the estimation of normalization statistics from downstream task demonstrations through a one-time foresight calibration. Prior to continual adaptation, FAN uses only two sources of prior information: coarse manipulation scenarios anticipated for the deployment domain and the capabilities of the target embodiment. No task descriptions, task-specific objects, demonstrations, or stream order are used at this stage.

For each scenario $\sigma$ in the anticipated set $\Sigma_{\mathrm{ant}}$, we manually decompose the manipulation process at changes in hand--object or object--object contact and characterize each phase by motion type, contact mode, gripper operation, and arm participation. Phases with similar execution requirements define motion primitives $\mathcal{P}_{\sigma}$, while the associated spatial and kinematic requirements define a reusable motion range $\mathcal{R}_{\sigma}$. Aggregating across scenarios yields
$
\mathcal{P}
=
\bigcup_{\sigma\in\Sigma_{\mathrm{ant}}}\mathcal{P}_{\sigma}$, $
\mathcal{R}
=
\bigcup_{\sigma\in\Sigma_{\mathrm{ant}}}\mathcal{R}_{\sigma}.
$
In our implementation, $\mathcal{P}$ includes reach, grasp,
push, pick, place, and rotate, while $\mathcal{R}$ spans
workspace regions and heights, motion directions, wrist
motions, gripper operations, and single-arm and bimanual
coordination patterns. This abstraction forms a compact
calibration checklist rather than an exhaustive manipulation
ontology.

The aggregated checklist guides collection of the calibration set
$\mathcal{C}
=
\{\tau_i^{\mathrm{cal}}\}_{i=1}^{N_{\mathcal{C}}}$.
In our experiments, $N_{\mathcal{C}}=8$ teleoperated trajectories are collected on the target platform. Each trajectory combines multiple primitives, and the set collectively covers $\mathcal{R}$. The trajectories contain neither downstream task instructions nor task-specific objects and are not demonstrations of benchmark tasks. The completed set is fixed before the task stream is revealed.

FAN estimates $\bm{q}(\mathcal{C})=
\left\{
\bm{q}_{0.01}(\mathcal{C}),
\bm{q}_{0.99}(\mathcal{C})
\right\}$ once from the calibration set and freezes it throughout continual adaptation. At every stage, these statistics normalize current-task and replay action targets and denormalize predicted actions during inference. Formally, 
$
\bm{q}_{\mathrm{train}}^{(k)}
=
\bm{q}_{\mathrm{replay}}^{(k)}
=
\bm{q}_{\mathrm{eval}}^{(k)}
=
\bm{q}(\mathcal{C})
$. The fixed coordinate system avoids task-specific replay re-encoding and keeps the numerical interface between the policy and robot unchanged across stages.

Freezing $\bm{q}(\mathcal{C})$ maintains Consistency; collecting $\mathcal{C}$ to cover the aggregated motion range targets Coverage; and completing calibration without downstream task data preserves Causality. FAN otherwise leaves the continual-learning pipeline unchanged.

\section{Experiments}\label{sec:experiments}

We conduct comprehensive real-robot experiments across four continual manipulation streams to isolate the effects of action normalization on task acquisition, transfer, and retention. These experiments further assess whether FAN enables robust adaptation across single-arm and bimanual settings with different task orders. Accordingly, the evaluation is structured around the following research questions:
\begin{itemize}
    \item \textbf{Q1:} How do different normalization strategies
    behave during continual VLA adaptation?
    \item \textbf{Q2:} Does FAN enable robust continual VLA adaptation across single-arm and bimanual task streams?
    \item \textbf{Q3:} How does task order affect continual VLA performance under different normalization strategies?
    \item \textbf{Q4:} How does the scope of a fixed normalization range affect subsequent-task acquisition?
\end{itemize}

\subsection{Experimental Setup}
\begin{figure}[h!]
    \centering
    \includegraphics[width=\linewidth]
    {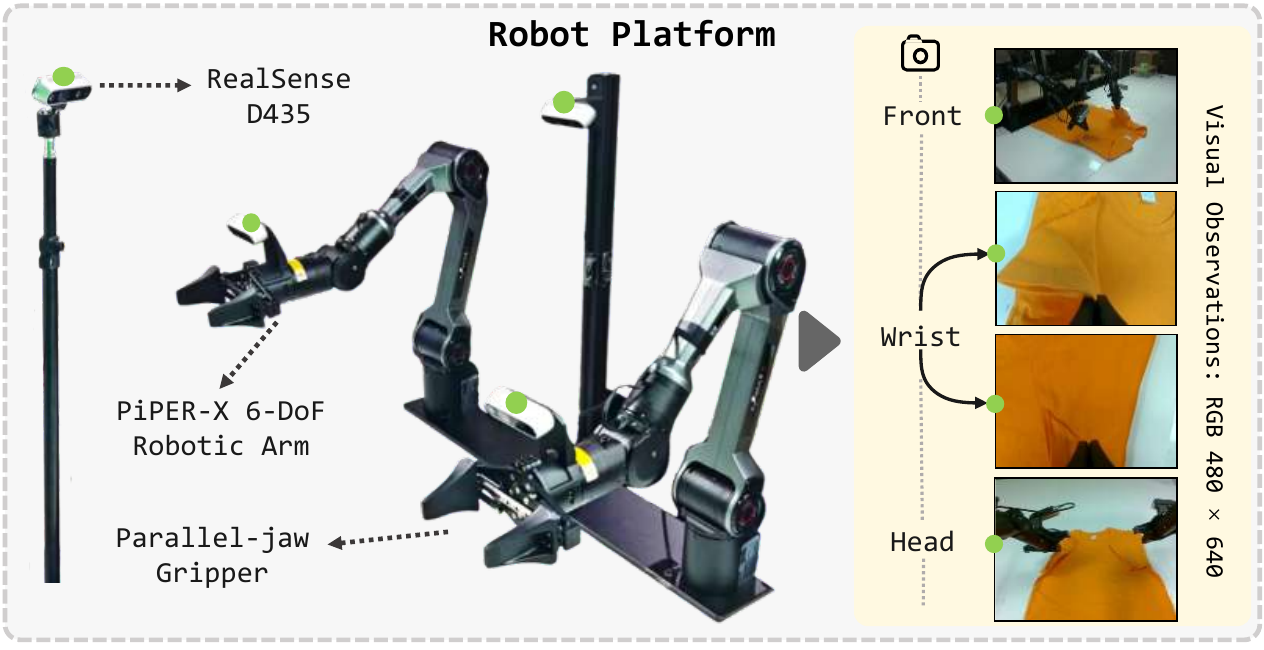}
    \caption{Robot platform and multi-view visual observations used during both data collection and policy deployment.}
    \label{fig:robot_setup}
    \vspace{-7pt}
\end{figure}

\begin{figure*}[t!]
    \centering
    \includegraphics[width=\textwidth]
    {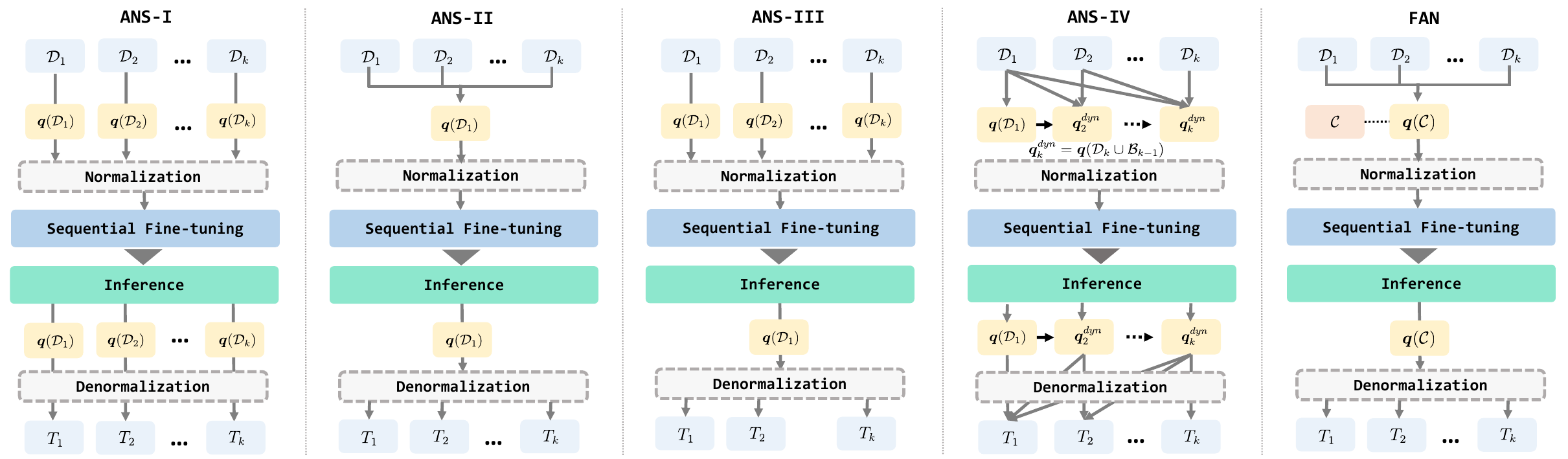}
    \caption{
    Action normalization strategies during continual fine-tuning and inference. Unlike the four task-derived baselines, FAN estimates $\bm{q}(\mathcal{C})$ solely from a task-independent calibration set and keeps it fixed throughout the stream.
    }
    \label{fig:normalization_strategies}
    \vspace*{-7pt}
\end{figure*}
\subsubsection{Robot Platform}
Fig.~\ref{fig:robot_setup} shows our bimanual platform, comprising two PiPER-X 6-DoF arms with parallel-jaw grippers and four RealSense D435 cameras (front, head, and one per wrist) that provide $480\times640$ RGB observations at 30 Hz. The policy predicts a 14-dimensional hybrid action with six delta joint commands and one absolute gripper target per arm:
$[\Delta\theta_1^L,\ldots,\Delta\theta_6^L,g^L,
\Delta\theta_1^R,\ldots,\Delta\theta_6^R,g^R]$.
For single-arm tasks, the seven left-arm action dimensions are clamped to zero. All methods share the same visual preprocessing and differ only in normalization protocol.

\subsubsection{Benchmark Design}
\begin{figure*}[!t]
    \centering
    \includegraphics[width=\textwidth]{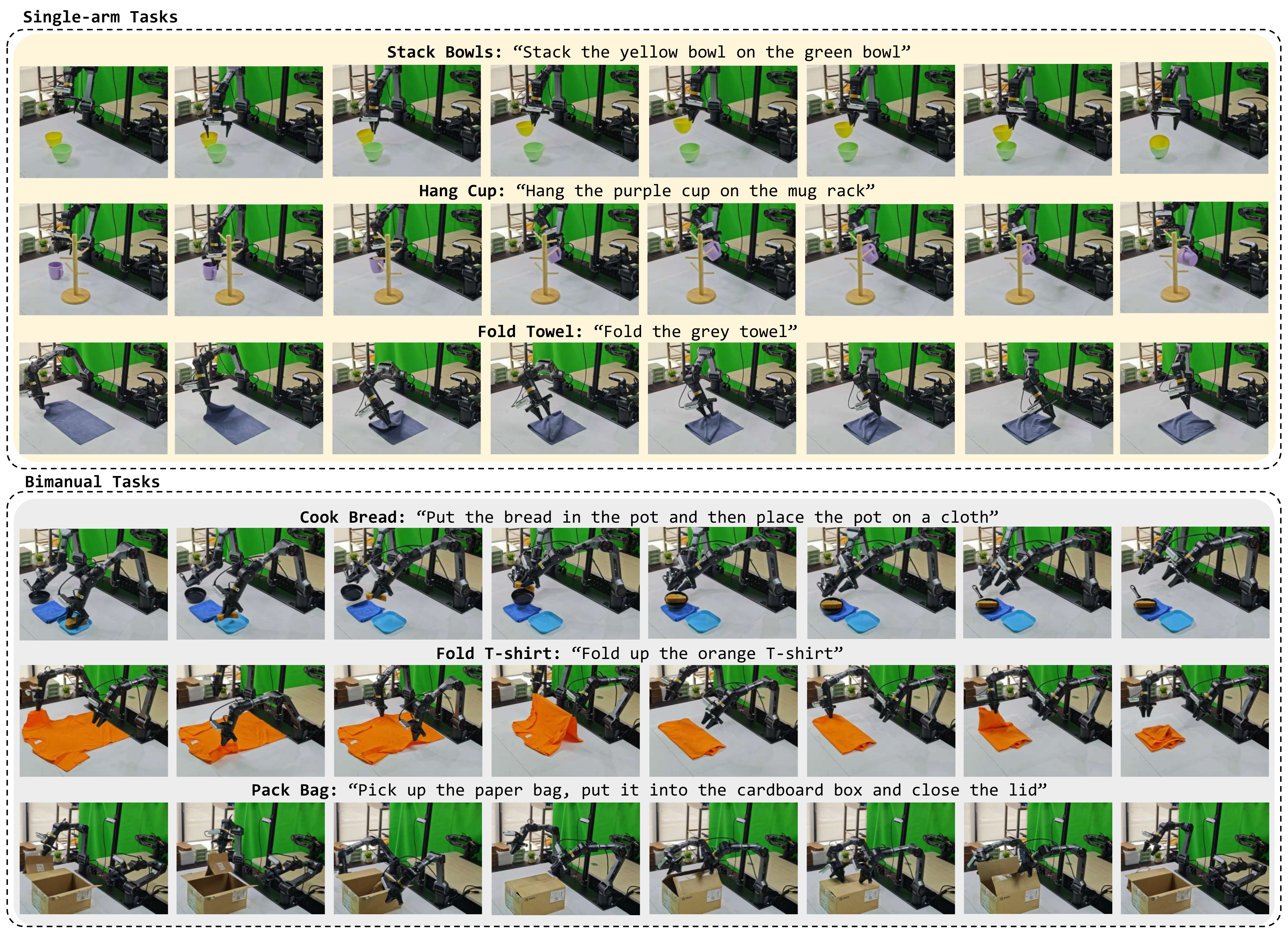}
   \caption{
   Representative rollout keyframes obtained with the final FAN checkpoint show physical robot execution on selected single-arm and bimanual tasks.
   }
    \label{fig:deployment-rollouts}
    \vspace*{-10pt}
\end{figure*}

Our benchmark comprises five single-arm and five bimanual tasks spanning rigid-object manipulation, contact-rich interaction, long-horizon execution, deformable-object manipulation, and bimanual coordination. To isolate sensitivity to task order, we evaluate two task orders for each embodiment:
\begin{itemize}
    \setlength{\itemsep}{0pt}
    \setlength{\parskip}{0pt}
    \setlength{\parsep}{0pt}

    \item \textbf{Single-Arm Stream~1:}
    \emph{Stack Bowls} $\rightarrow$ \emph{Hang Cup}
    $\rightarrow$ \emph{Press Button} $\rightarrow$
    \emph{Fold Towel} $\rightarrow$ \emph{Push Box}.

    \item \textbf{Single-Arm Stream~2:}
    \emph{Press Button} $\rightarrow$ \emph{Fold Towel}
    $\rightarrow$ \emph{Stack Bowls} $\rightarrow$
    \emph{Push Box} $\rightarrow$ \emph{Hang Cup}.

    \item \textbf{Bimanual Stream~1:}
    \emph{Cook Bread} $\rightarrow$ \emph{Place Cola}
    $\rightarrow$ \emph{Place Fruits} $\rightarrow$
    \emph{Fold T-shirt} $\rightarrow$ \emph{Pack Bag}.

    \item \textbf{Bimanual Stream~2:}
    \emph{Place Cola} $\rightarrow$ \emph{Place Fruits}
    $\rightarrow$ \emph{Pack Bag} $\rightarrow$
    \emph{Fold T-shirt} $\rightarrow$ \emph{Cook Bread}.
\end{itemize}
Fig.~\ref{fig:deployment-rollouts} shows examples of
real-robot policy rollouts from Single-Arm and Bimanual Stream 1.

Across all normalization strategies, we use experience replay (ER) as the continual-learning procedure and sequentially fine-tune the base VLA policy $\pi_{0.5}$ over each task stream. At stage $k>1$, the replay buffer $\mathcal{B}_{k-1}$ stores episodes from all previous tasks. Its total capacity is set to $0.2$ of the per-task demonstration count and divided uniformly among the $k-1$ tasks. At each training step, a mini-batch is drawn either from $\mathcal{B}_{k-1}$ with probability $f_r=0.2$ or from $\mathcal{D}_k$ with probability $1-f_r=0.8$.

\subsubsection{Normalization Strategies}

Following Sec.~\ref{subsec:action_normalization}, $\bm{q}(\mathcal{D})$ denote the per-dimension 1st- and 99th-percentile statistics estimated from dataset $\mathcal{D}$. These statistics are used to normalize action targets during training and to denormalize the policy's normalized action predictions during inference. Figure~\ref{fig:normalization_strategies} summarizes the five resulting protocols, whose stage-wise definitions are detailed below:

\begin{itemize}
\item \textbf{ANS-I} applies $\bm{q}(\mathcal{D}_j)$ to task $j$
    throughout training, replay, and evaluation.

\item \textbf{ANS-II} applies $\bm{q}(\mathcal{D}_1)$ throughout training, replay, and evaluation at every stage.

\item \textbf{ANS-III} applies task-specific $\bm{q}(\mathcal{D}_j)$ during training and replay but evaluates every task with $\bm{q}(\mathcal{D}_1)$.

\item \textbf{ANS-IV} recomputes $\bm{q}^{\mathrm{dyn}}_k=\bm{q}(\mathcal{D}_k\cup\mathcal{B}_{k-1})$ and applies it to current-task data, replay data, and all evaluations at stage $k$.

\item \textbf{FAN} applies $\bm{q}(\mathcal{C})$, defined in
 Sec.~\ref{subsec:fan}, to current-task, replay, and evaluation samples at every stage.
\end{itemize}

\subsubsection{Evaluation Metrics}
Each rollout receives a normalized score in $[0,100]$ under a fixed, task-specific stepwise rubric.
For each task, all methods and checkpoints use the same rubric, initialization procedure, environment perturbation range, and evaluation protocol. Each  checkpoint--task score is the mean over 10 real-robot rollouts from the corresponding trained checkpoint.

Let $c_{i,j}$ denote the score on task $T_j$ after training through task $T_i$, where $i\geq j$, and let $K=5$. The average score at the final checkpoint is defined as $\mathrm{AS}=\frac{1}{K}\sum_{j=1}^{K}c_{K,j}$. 
Backward transfer measures the final change on previously
learned tasks:
 $\mathrm{BWT}=\frac{1}{K-1} \sum_{j=1}^{K-1}
 \left(c_{K,j}-c_{j,j}\right)$. Higher BWT indicates better retention or beneficial backward transfer, while negative BWT indicates forgetting. Forward transfer measures whether previously learned tasks facilitate the acquisition of subsequent tasks:
$ \mathrm{FWT}=\frac{1}{K-1}  \sum_{j=2}^{K}
 \left(c_{j,j}-c_j^{\mathrm{ST}}\right)$,
where $c_j^{\mathrm{ST}}$ is obtained by fine-tuning on task $j$ alone using that task's own normalization statistics. Task~1 is excluded because it has no preceding task from which forward transfer can arise. 

\begin{table}[t]
\centering
\caption{Stream-level and cross-stream performance under
different action-normalization strategies.}
\label{tab:overall_performance}

\footnotesize
\renewcommand{\arraystretch}{1.03}
\setlength{\tabcolsep}{2pt}

\begin{tabular*}{\columnwidth}{@{\extracolsep{\fill}}lrrrrr@{}}
\toprule
\multicolumn{6}{l}{\textbf{(a) Average Score (AS, $\uparrow$)}} \\
\cmidrule(lr){1-6}
Method & SA-1 & SA-2 & BI-1 & BI-2 & Avg. $\uparrow$ \\
\midrule
ANS-I
& 74.2 & 78.5 & 53.0 & 42.3 & 62.0 \\
ANS-II
& 97.2 & 45.7 & \textbf{97.2} & 68.0 & 77.0\\
ANS-III
& 51.3 & 22.0 & 17.3 & 13.0 & 25.9\\
ANS-IV
& 73.7 & 73.0 & 56.8 & 32.5 & 59.0\\
\textbf{FAN (ours)}
& \textbf{98.4}
& \textbf{94.4}
& 95.7
& \textbf{92.7}
& \textbf{95.3} \\

\midrule
\multicolumn{6}{l}{\textbf{(b) Backward Transfer (BWT, $\uparrow$)}} \\
\cmidrule(lr){1-6}
Method & SA-1 & SA-2 & BI-1 & BI-2 & Avg. $\uparrow$ \\
\midrule
ANS-I
& -4.8 & -15.1 & \textbf{10.2} & -15.4 & -6.3  \\
ANS-II
& 1.5 & \textbf{0.1} & 1.9 & \textbf{-2.1} & \textbf{0.4} \\
ANS-III
& \textbf{2.6} & 0.0 & -1.7 & -6.3 & -1.4  \\
ANS-IV & -13.1 & -21.2 & -20.2 & -36.4 &-22.7 \\
\textbf{FAN (ours)}
& 2.3 & -1.3 & 4.4 & -5.0 & 0.1 \\

\midrule
\multicolumn{6}{l}{\textbf{(c) Forward Transfer (FWT, $\uparrow$)}} \\
\cmidrule(lr){1-6}
Method & SA-1 & SA-2 & BI-1 & BI-2 & Avg. $\uparrow$ \\
\midrule
ANS-I
& -11.1 & 4.6 & -54.0 & -41.7 & -25.6\\
ANS-II
& 11.4 & -51.6 & \textbf{9.6} & -22.9 & -13.4\\
ANS-III
& -47.1 & -81.1 & -86.7 & -87.5 & -75.6 \\
ANS-IV
& -3.5 & 3.9 & -18.8 & -32.5 & -12.7 \\
\textbf{FAN (ours)}
& \textbf{12.1}
& \textbf{10.6}
& 5.2
& \textbf{8.3}
& \textbf{9.1} \\
\bottomrule
\vspace{-10pt}
\end{tabular*}
\end{table}

\subsection{Overall Comparison of Action Normalization Strategies}
\label{sec:overall_results}

Table~\ref{tab:overall_performance} compares the five normalization strategies across two single-arm (SA) and two bimanual (BI) streams; Avg. denotes the mean over the four streams.

ANS-I achieves an average AS of 62.0. Its average
FWT of $-25.6$ indicates limited transfer across tasks.
Thus, although task-specific statistics provide suitable
scaling for each individual task, the resulting heterogeneous
coordinates impede knowledge reuse within a shared policy.
ANS-II is the strongest baseline on average, with an AS of 77.0 and a BWT of 0.4. 
Its fixed coordinate system performs well in both Stream 1 settings but yields markedly lower AS and negative FWT on the two reordered streams.

This contrast shows that fixed coordinates alone do not ensure robust continual adaptation across task orders.
ANS-III has the lowest AS (25.9) and FWT ($-75.6$), consistent with the train--test coordinate mismatch. Its average BWT of $-1.4$ is therefore not evidence of effective retention because later tasks are weakly acquired.
ANS-IV obtains an average AS of $59.0$. It produces negative BWT on all four streams, averaging -22.7, indicating that dynamically updating normalization range comes at a substantial retention cost. 

These results answer Q1: the source, update schedule, and application of normalization statistics jointly shape task acquisition, cross-task transfer, and retention, producing distinct failure modes.

\subsection{FAN Enables Robust Continual Adaptation}

Table~\ref{tab:overall_performance} shows that FAN achieves the highest average AS (95.3) and the smallest cross-stream AS spread (92.7 to 98.4) among the compared strategies. Its positive FWT on all four streams indicates that subsequent-task acquisition scores exceed the common single-task reference
on average within each stream. Its near-zero average BWT (0.1) indicates little net change in performance on previously learned tasks. 
These results answer Q2: FAN enables robust continual VLA learning across the evaluated single-arm and bimanual streams, with a favorable balance between new-task acquisition and retention.

\begin{figure}[t!]
  \centering
  \includegraphics[width=\columnwidth]
  {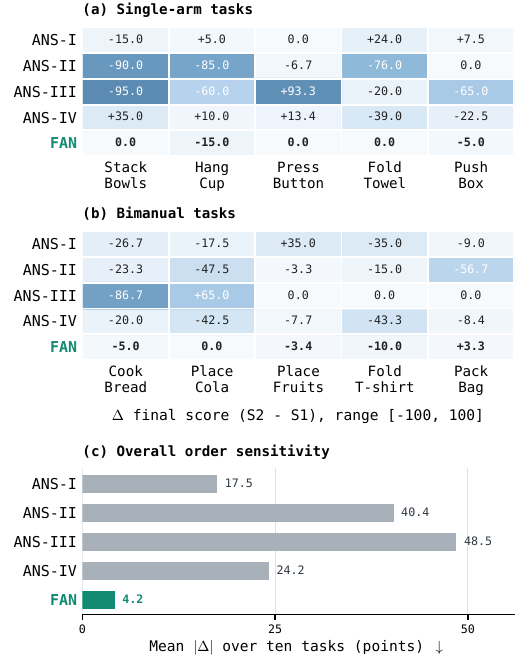}
  \caption{Task-order sensitivity after matching tasks by identity. (a)--(b) Signed final-score differences $\Delta$ for the single-arm and bimanual tasks, computed as the score in Stream 2 minus that in Stream 1. $\Delta>0$ and $\Delta<0$ indicate higher final performance in Stream 2 and Stream 1, respectively. (c) Mean absolute difference across all ten matched tasks.
  }
  \label{fig:task_order_sensitivity}
  \vspace{-8pt}
\end{figure}

\subsection{Task-Order Sensitivity across Normalization Strategies}
\label{sec:task_order_sensitivity}

Each stream pair contains the same tasks in different orders, allowing task-order effects to be assessed while holding task identity fixed. For each normalization strategy, we match tasks by identity and compute the signed final-score difference for each matched task as the Stream 2 score minus the Stream 1 score. Overall task-order sensitivity is summarized by the mean absolute difference across the ten matched tasks. Fig.~\ref{fig:task_order_sensitivity} reports the task-level signed differences and the resulting mean absolute difference.

Across the baselines, the mean absolute difference ranges from 17.5 to 48.5 points. ANS-II exhibits a directional effect: when \textit{Press Button} and \textit{Place Cola} are first in Stream 2, all matched-task differences are non-positive [Fig.~\ref{fig:task_order_sensitivity}(a)--(b)], consistent with first-task-dependent coverage. ANS-III is the most sensitive overall, while ANS-I remains moderately sensitive. For ANS-IV, each order produces a different sequence of updates to statistic and hence an order-dependent trajectory of coordinate drift.

FAN reduces the mean absolute difference to 4.2 points, with all task-level differences between $-15.0$ and $+3.3$ points [Fig.~\ref{fig:task_order_sensitivity}(a)--(c)]. These results answer Q3: task order has a larger effect when normalization coordinates depend on individual tasks, the first task, or an order-dependent update sequence, whereas FAN's fixed, task-independent coordinates substantially reduce sensitivity across the evaluated orderings.

\begin{figure}[t]
  \centering
  \includegraphics[width=\columnwidth]
  {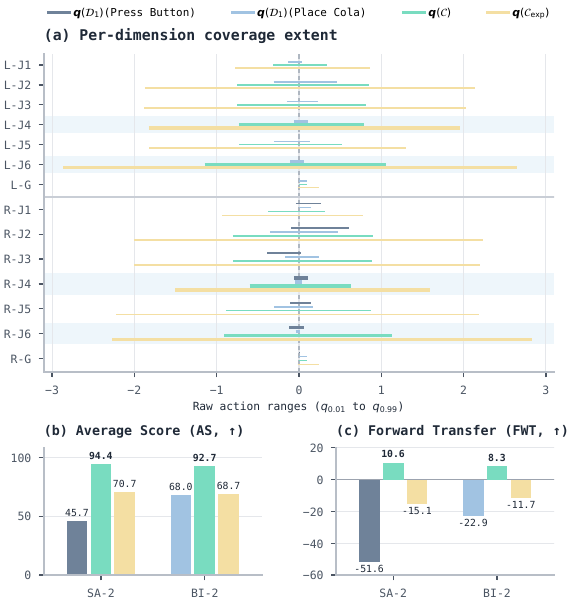}
  \caption{Coverage ablation with fixed normalization statistics on SA-2 and BI-2. (a) Per-dimension raw-action intervals $[q_{0.01,d}(\mathcal{D}),q_{0.99,d}(\mathcal{D})]$ for $\mathcal{D}\in \{\mathcal{D}1,\mathcal{C},\mathcal{C}{\mathrm{exp}}\}$. For $\bm{q}(\mathcal{D}_1)$ on SA-2, only the seven active right-arm dimensions are shown. (b)--(c) AS and FWT for the three fixed normalization ranges. Across the evaluated streams, the $\bm{q}(\mathcal{C})$ condition outperforms the narrower and broader alternatives in both overall performance and subsequent-task acquisition.}
  
  \label{fig:coverage_ablation}
  \vspace{-7pt}
\end{figure}

\subsection{Effect of Normalization Range on Task Acquisition}

Fig.~\ref{fig:coverage_ablation} isolates normalization-range coverage by comparing 
$\bm{q}(\mathcal{D}_1)$, $\bm{q}(\mathcal{C})$, and
$\bm{q}(\mathcal{C}_{\mathrm{exp}})$. Here, $\mathcal{D}_1$ denotes the \textit{Press Button} dataset in SA-2 and the \textit{Place Cola} dataset in BI-2, while $\bm q(\mathcal C_{\mathrm{exp}})$ is derived directly from $\bm q(\mathcal C)$ by 
preserving the interval midpoint while expanding its percentile width in every dimension by a factor of 2.5.
All three sets of statistics remain fixed throughout training, replay, and evaluation, excluding inter-stage coordinate drift.

Compared with $\bm{q}(\mathcal{C})$, $\bm{q}(\mathcal{D}_1)$ spans narrower ranges in several action dimensions [Fig.~\ref{fig:coverage_ablation}(a)], amplifying normalized variations by up to $12.0\times$ in SA-2 and $34.7\times$ in BI-2. Under $\bm{q}(\mathcal{D}_1)$, both streams exhibit lower AS and negative FWT [Fig.~\ref{fig:coverage_ablation}(b)--(c)]. This degradation is consistent with percentile mapping, which scales a given physical variation inversely with range width and therefore turns later motions into disproportionately large normalized targets.
At the other extreme, $\bm{q}(\mathcal{C}_{\mathrm{exp}})$ spans broader ranges than $\bm{q}(\mathcal{C})$ in several action dimensions [Fig.~\ref{fig:coverage_ablation}(a)]. Replacing $\bm{q}(\mathcal{C})$ with $\bm{q}(\mathcal{C}_{\mathrm{exp}})$ reduces AS by 23.7 and 24.0 points and FWT by 25.7 and 20.0 points on SA-2 and BI-2, respectively, making FWT negative in both streams [Fig.~\ref{fig:coverage_ablation}(b)--(c)]. Broader ranges compress task-relevant physical variations in normalized coordinates and map a given normalized prediction error to a larger physical error after denormalization, consistent with the observed degradation.

Of the three fixed-statistics conditions, $\bm{q}(\mathcal{C})$ yields the highest AS and FWT on both streams, indicating sufficient coverage without excessive compression. 
These results answer Q4: with statistics fixed, both insufficient and overly broad coverage can impair subsequent-task acquisition. On the evaluated platform, the normalization range should therefore match the anticipated embodiment-level motion requirements.

\section{Conclusions}

This work shows that action normalization is a consequential design choice in replay-based continual VLA adaptation. Across four real-robot streams, the evaluated protocols exhibit distinct failure modes: task-dependent coordinates limit knowledge reuse, first-task statistics can under-cover subsequent motions, train--test mismatches impair deployment, and online updates introduce inter-stage coordinate drift. These findings motivate the three requirements of Consistency, Coverage, and Causality.

FAN satisfies these requirements by estimating normalization statistics once from a small, task-independent calibration set and freezing them across sequential fine-tuning and inference. It achieves the highest average AS of 95.3, positive FWT on all four streams, and near-zero average BWT. The fixed-statistics coverage ablation further shows that both narrow task-derived ranges and unnecessarily broad ranges impair subsequent-task acquisition, indicating that coverage should be matched to anticipated embodiment-level motion requirements rather than maximized. FAN also yields the lowest matched-task variation across the evaluated stream orders. Together, these results demonstrate that a fixed, causally available, and appropriately scoped action coordinate system provides a simple and effective basis for continual VLA adaptation.

\section*{Acknowledgements}
This work is supported, in part, by the HK UGC Early Career Scheme under Grant No. HKU 27205826, and the National Natural Science Foundation of China under Grant No. 62603568. We also thank INFIFORCE for providing the computing resources.


\bibliographystyle{IEEEtran}
\bibliography{ref}

\end{document}